\documentclass[a4paper]{article}

\usepackage{amsmath,amssymb,amsthm}
\usepackage{graphicx}

\usepackage{bm}

\usepackage{url}
\usepackage{algorithm}
\usepackage{algorithmic}
\usepackage{enumitem}
\usepackage[subrefformat=parens]{subcaption}
\usepackage{ifpdf}

\newcommand{\R}{\mathbb{R}}

\usepackage{pgfplots}
\pgfplotsset{compat=1.12}

\DeclareMathOperator*{\argmin}{arg\,min}
\DeclareMathOperator*{\argmax}{arg\,max}

\newcommand{\cvp}{\stackrel{P}{\rightarrow}}

\newtheorem{theorem}{Theorem}[section]
\newtheorem{lemma}{Lemma}[section]

\newtheorem{assumption}{Assumption}[section]
\newtheorem{result}{Result}[section]

\title{On Estimation of Conditional Modes\\ Using Multiple Quantile Regressions}

\author{Hirofumi Ohta \\ The University of Tokyo, \\Graduate School of Economics,\\ Department of Statistics.\\JST, ERATO,\\ Kawarabayashi Large Graph Project. \\ \url{hirofumi-ohta@g.ecc.u-tokyo.ac.jp} \and Satoshi Hara\\ Osaka University, Japan\\ \url{satohara@ar.sanken.osaka-u.ac.jp}}
\begin{document}
\maketitle
\begin{abstract}
	We propose an estimation method for the conditional mode when the conditioning variable is high-dimensional.
	In the proposed method, we first estimate the conditional density by solving quantile regressions multiple times.
	We then estimate the conditional mode by finding the maximum of the estimated conditional density.
	The proposed method has two advantages in that it is computationally stable because it has no initial parameter dependencies, and it is statistically efficient with a fast convergence rate.
	Synthetic and real-world data experiments demonstrate the better performance of the proposed method compared to other existing ones.
\end{abstract}

\section{Introduction}

The estimation of the conditional mode, or \textit{modal regression}~\cite{yao2014new,kemp2012quantile,chen2016nonparametric,sasaki2016modal}, is an important topic in statistics~\cite{sager1982maximum, yao2012local,yao2014new}, econometrics~\cite{lee1989mode,lee1993quadratic, damien2017bayesian, krief2017semi,kemp2012quantile}, and machine learning~\cite{feng2017statistical,sasaki2016modal}.
Compared to ordinary regression, modal regression is particularly useful when the data distribution is highly skewed and has fat tails.
In such a situation, ordinary regression, which estimates the conditional mean of the distribution, fails to capture the major trend underlying the data.
This is because the conditional mean is not necessarily the point where the data points distribute densely, i.e., it can be far away from the majority of the data. 
Conditional mode is a convenient alternative to the conditional mean in this situation as it can capture the majority of the data. 
Hence, with modal regression, we can find a major trend underlying the data.
For example in economics, modal regression is shown to be useful when analyzing the relationship between GDP and several quantities such as CO2~\cite{huang2012mixture} and the stock index~\cite{huang2013nonparametric}. This is because these quantities have highly skewed distributions and ordinary regression methods cannot capture the major trend of the data.
Not only in economics, but modal regression is used also in several data analysis tasks such as the traffic data~\cite{einbeck2006modelling} and the forest fire data~\cite{yao2014new}.

Modal regression has been studied primarily in statistical literatures.
In particular, several studies have proposed modal regression methods based on the kernel density estimation~\cite{yao2014new,kemp2012quantile,chen2016nonparametric,sasaki2016modal}.
For example, Yao and Li~\cite{yao2014new} and Kemp and Santos-Silva~\cite{kemp2012quantile} proposed a linear modal regression method based on the kernel density estimation.
Chen et al.~\cite{chen2016nonparametric} proposed modeling the conditional density function directly using the kernel density function.
The modal regression problem is then formulated as finding the maximizer of the estimated conditional density function.
The approach of Sasaki et al.~\cite{sasaki2016modal} is similar to Chen's method, while the authors proposed using a log-density derivative estimation.

While several modal regression methods have been proposed, there still remain two challenges.
The first challenge is the computational instability.
All of the existing methods rely on non-convex optimization problems, and none of the methods have a global convergence guarantee.
This means that these methods can stack	 local optima and fail to find the conditional mode.
This undesirable property leads to an unstable conditional mode estimation where the estimated mode can vary depending on the initial parameters.	
The second challenge is the statistical inefficiency.
Some of these methods have a very slow convergence rate, which is a exponential of the input dimension $p$.
This causes the well-known curse of dimensionality where the estimator is no longer useful if the input dimension $p$ is large.

In this study, we propose a new modal regression method that can overcome these two challenges.
The proposed method is based on the quantile regression~\cite{koenker1987estimation,koenker2005quantile}.
In the method, instead of the commonly used kernel density estimation, we use quantile regression to model the conditional density function.

Our major contributions are twofold.
First, we propose a new modal regression method based on the quantile regression. 
The advantage of the proposed method is that all the steps in the method are convex optimization problems and there is no dependency on the initial parameters.
Therefore, the proposed method is computationally very stable unlike existing methods.

Second, we show that the proposed method is statistically efficient in that it has a faster convergence rate than the existing methods.
Specifically, we show that the error rate of the proposed method is not exponential of the input dimension $p$.
This implies that the proposed method can avoid the curse of dimensionality and scale well with high-dimensional data.

Our experimental results confirm that the proposed method can indeed overcome the two above-mentioned challenges.
We found that the proposed method can obtain smaller test mean square errors than existing methods.
The results also indicate that the proposed method is computationally stable, as expected.

This paper is organized as follows: 
In Section~\ref{sec:quantile}, we briefly introduce quantile regression which forms the basis of this study. 
We provide the proposed modal regression method in Section~\ref{sec:proposed}, and we study its theoretical properties in Section~\ref{sec:theory}.
In Section~\ref{sec:compare}, we discuss our theoretical results and compare them to other existing studies.
Section~\ref{sec:extension} shows some possible extensions of the proposed method.
In Section~\ref{sec:experiment}, we demonstrate our experimental results on synthetic and real-world datasets, showing the advantages of the proposed method.
Finally, we conclude the paper in Section~\ref{sec:conclusion}.

\paragraph{Notation}
Let $Y\in \R$ be scalar and $X\in \R^p$ be vector random variables. 
We denote the conditional density of $Y$ given $X$, or in short, $Y|X$, as $f(y|x)$.
For any $\tau \in (0, 1)$, we denote the $\tau$-th quantile function of $Y$ given $X=x$ by $Q_{Y|X}(\tau|x)$.
$Z_n \cvp Z$ indicates that $Z_n$ converges in probability toward $Z$. 
For a random variable $Z$, $Z \sim$ U$ ([a, b]^p) $ indicates that $Z$ is uniformly distributed with the support $[a,b]^p$. 
$Z \sim$ Ga$(k, \theta)$ indicates that $Z$ is gamma-distributed with a shape parameter $k$ and a scale parameter $\theta$. 
$Z \sim$ N$(\mu, \sigma^2)$ indicates that $Z$ is normally distributed with a mean $\mu$ and a variance $\sigma^2$.

\section{Preliminaries}
\label{sec:quantile}

In this section, we briefly review linear quantile regression~\cite{koenker1987estimation,koenker2005quantile} which constitutes the basis of our study.

\subsection{Quantile Regression}

In linear quantile regression, for any $\tau \in (0, 1)$, we model the $\tau$-th quantile function $Q_{Y|X}(\tau|x)$ as a linear function:
\begin{align}
Q_{Y|X}(\tau | x) := x^\top \beta(\tau) ,
\label{eq:quantile}
\end{align}
where $\beta(\tau) \in \R^p$ is a regression coefficient that depends on $\tau$.

The regression coefficient $\beta(\tau)$ can be derived by solving a linear programming~\cite{koenker1987estimation}.
Suppose that we observed $n$ independent and
identically distributed sample $\{x_i, y_i\}_{i=1}^n$ from the joint density of $x$ and $y$.
Then, the estimation of $\beta(\tau)$ can be formulated as the next optimization problem~\cite{koenker1987estimation}:
\begin{align}
\hat{\beta}(\tau) := \argmin_{b \in \R^p} \sum_{n=1}^n \rho_\tau(y_i - x_i^\top b), 
\label{eq:quantile_opt}
\end{align}
where $\rho_\tau(u) = \tau$ if $u \ge 0$ and $\rho_\tau(u) = 1-\tau$ otherwise.
This problem can be reformulated as the following linear programming:
\begin{align}
& \min_{b, v, w} \sum_{i=1}^n \left( \tau v_i + (1 - \tau) w_i \right), \nonumber \\
& {\rm s.t.} \; \forall i, x_i^\top b + v_i - w_i = y_i, {\rm and} \; v_i, w_i \ge 0.
\end{align}
This problem can be solved efficiently, e.g., by using the interior point method~\cite{portnoy1997gaussian}.
We denote the estimated $\tau$-th quantile by $\hat{Q}_{Y|X}(\tau|x) := x^\top \hat{\beta}(\tau)$.

\subsection{Statistical Property of Quantile Regression}

For the regression coefficient estimator $\hat{\beta}(\tau)$, it is known that $\sqrt{n}\left( \hat{\beta}(\tau) - \beta(\tau) \right) = O_p(1)$ holds for any $\tau \in \mathcal{T}$ where $\mathcal{T}$ is any compact subset of $(0, 1)$.
\begin{lemma}[Koenker and Portnoy~\cite{koenker1987estimation}]
	Let $\mathcal{T}= [\varepsilon, 1-\varepsilon]$ with $\varepsilon \in (0, 1/2)$.
	Under the regularity conditions of Koenker and Portnoy~\cite{koenker1987estimation} and Portnoy~\cite{portnoy1984tightness}, the following equation holds:
	\begin{align}
	\sup_{\tau \in \mathcal{T}} \left\| \hat{\beta}(\tau) - \beta(\tau) \right\| = O_p\left(\frac{1}{\sqrt{n}}\right).
	\end{align}
	\label{lem:quantile}
\end{lemma}
This lemma ensures that the linear quantile regression estimator converges at a parametric rate, except at the edge of the quantile index set.

\section{Proposed Modal Regression Method}
\label{sec:proposed}

Typical approach to the modal regression problem consists of the next two steps.
First, we estimate the conditional density $f(y|x)$ from the data $\{x_i, y_i\}_{i=1}^n$.
Second, we estimate the conditional mode as $\argmax_y f(y|x)$.
The technical challenge is, therefore, how to accurately estimate the conditional density $f(y|x)$.
Here, we propose estimating the conditional density $f(y|x)$ using quantile regression.		
Following Koenker~\cite{koenker2005quantile}, by using the chain rule of differentiation, 
\begin{align}
\frac{\partial Q_{Y|X}(\tau|x)}{\partial \tau} = \frac{1}{f(Q_{Y|X}(\tau | x)|x)}
\end{align}
holds. Then, the conditional density is approximated as below:
\begin{align}
f(Q_{Y|X}(\tau | x)|x) &= \left( \frac{\partial Q_{Y|X}(\tau|x)}{\partial \tau} \right)^{-1} \nonumber \\
& \approx \frac{2h}{Q_{Y|X}(\tau + h | x) - Q_{Y|X}(\tau - h | x)},
\label{eq:truecond}
\end{align}
where we replaced the derivative with the difference in the second line.
Therefore, we can estimate the conditional density function as
\begin{align}
\hat{f}(Q_{Y|X}(\tau | x)|x) = \frac{2h}{\hat{Q}_{Y|X}(\tau + h | x) - \hat{Q}_{Y|X}(\tau - h | x)}, 
\label{eq:cond}
\end{align}

Here, $\hat{Q}_{Y|X}(\tau|x)$ is the estimated $\tau$-th quantile function, and $h$ is a bandwidth satisfying $h \to 0$ as $n \to \infty$.
This estimator, Eq.\ (\ref{eq:cond}), has also been studied by Belloni et al.~\cite{belloni2016valid} and Bradic and Kolar~\cite{bradic2017uniform}.
Once the conditional density is estimated for several different values of $\tau$, we can derive the mode estimator as the quantile with the largest estimated conditional density.

Algorithm~\ref{alg:proposed} shows the proposed modal regression method based on the conditional density estimation using quantile regression.
In the algorithm, we estimate the conditional density for each value of $\tau$ in the candidate set $T$.
The algorithm then finds the optimal $\tau$ with the maximum conditional density and returns the corresponding quantile function estimate $\hat{Q}_{Y|X}(\tau|x)$ as the mode estimator.
We note that the larger number of candidates of $\tau$ in the set $T$ is always desirable because it leads to a better mode estimator.
This can however incur high-computational complexity.
Nevertheless, in our simulation study, we found that the algorithm is still computationally feasible even if we set the candidate size to $|T| = 1000$. 

\begin{figure}[h]
	\centering
	\begin{algorithm}[H]
		\small
		\caption{Modal Regression Algorithm}
		\label{alg:proposed}
		\begin{algorithmic}
			\REQUIRE Dataset $\{x_i, y_i\}_{i=1}^n$, Test point $x$, Candidates of quantile $T$, Bandwidth $h$
			\ENSURE Estimated conditional mode $\hat{m}(x)$
			\STATE $\hat{m}(x) \leftarrow 0$
			\STATE $\hat{f} \leftarrow 0$
			\FOR{$\tau \in T$}
			\STATE $\hat{f}' \leftarrow$ Estimated conditional density (\ref{eq:cond})
			\IF{$\hat{f}' > \hat{f}$}
			\STATE $\hat{m}(x) \leftarrow \hat{Q}_{Y|X}(\tau|x)$
			\STATE $\hat{f} \leftarrow \hat{f}'$
			\ENDIF
			\ENDFOR
		\end{algorithmic}
	\end{algorithm}
\end{figure}

Note that all the steps in Algorithm~\ref{alg:proposed} have no initial parameter dependencies.
This is because all the steps in the algorithm are composed of only the quantile regression and the argmax operation over the estimated conditional density.
Because the quantile regression can be solved using linear programming, its global optimality is guaranteed.
The argmax operation is applied in one dimension, and the naive search as in Algorithm~\ref{alg:proposed} can find the global optima.

\section{Theoretical Analyses}
\label{sec:theory}

In this section, we provide a consistency guarantee of the proposed mode estimator as well as its convergence rate.
Here, we use the following notations.
Let $\tau_m(x)$ be the quantile that gives the true conditional mode, i.e., 
\begin{align}
\tau_m(x) := \argmax_\tau f(Q_{Y|X}(\tau | x) | x).
\end{align}
We denote its estimator as
\begin{align}
\hat{\tau}_m(x) := \argmax_\tau \hat{f}(Q_{Y|X}(\tau | x) | x),
\end{align}
where $\hat{f}$ is the estimated conditional density defined in Eq.\ (\ref{eq:cond}).
We define the true mode and its estimator as follows:
\begin{align}
m(x) &:= Q_{Y|X}(\tau_m(x) | x) , \\
\hat{m}(x) &:= \hat{Q}_{Y|X}(\hat{\tau}_m(x) | x).
\end{align}
In this section, we only provide proof overviews of the theorems.
All the detailed proofs can be found in the supplemental material (Appendix).

\begin{assumption}
	In our analysis, we adopt the following assumptions.
	\begin{enumerate}[leftmargin=24pt]
		\item[\rm A0.] Quantile functions are linear, i.e., $Q_{Y|X}(\tau | x) := x^\top \beta(\tau)$, $\forall \tau \in (0,1)$.
		\item[\rm A1.] $\forall x \in \R^p, \forall y \in \R$, $\exists c > 0$ such that $0<f(y|x)\leq c$ and $\sup_{|\bar{\tau}-\tau| \leq h } |Q'''_{Y|X}(\bar{\tau}|x)| \leq c$.
		\item[\rm A2.] $\forall x \in \R^p$, $f(y|x)$ has a unique global mode and $\tau_m(x) \in \mathcal{T}$ for a compact set $\mathcal{T}$.
		\item[\rm A3.] $\forall x \in \R^p$, $f(y|x)$ is Lipschitz continuous with a constant $L_x$, i.e., $|f(y|x) - f(y'|x)| \le L_x |y - y'|$.
		\item[\rm A4.] Identification condition: $m(\cdot)$ satisfies that, for all $\varepsilon>0$ and any function $\hat{m}(\cdot)$, there exists $\delta>0$ such that $|\hat{m}(x) - m(x)| \geq \varepsilon$ implies $\left|f(\hat{m}(x)|x) - f(m(x)|x) \right| \geq \delta$.
		\item[\rm A5.] Polynomial majorant: There exist positive constants $\delta'$, $K$, $\gamma_1$, and $\gamma_2$ with
		$\gamma_1 \gamma_2 \geq 1$ such that, for any $\varepsilon' \in (0,1)$, there exist positive constants $K_{\varepsilon'}$ and $n_\varepsilon'$ such that for all $n \geq n_{\varepsilon'}$,
		\begin{align}
		& f(Q_{Y|X}(\tau|x)|x) \nonumber \\
		& \hspace{6pt} \leq \sup_{\tau \in \mathcal{T}} f(Q_{Y|X}(\tau|x)|x) - K(|\tau-\tau_m|\wedge \delta')^{\gamma_1} ,
		\end{align}
		holds uniformly on the set $\{ \tau \in \mathcal{T}: |\tau - \tau_m| \geq (K_{\varepsilon'}/a_n)^{\gamma_2}\}$ with a probability of  at least $1-\varepsilon'$, where $1/a_n = h^2  + (nh^2)^{-1/2}.$
	\end{enumerate}
\end{assumption}

Assumption A1 requires the conditional density and the third-order derivative of the quantile function to be upper-bounded.
This assumption is satisfied when the value of the quantile function does not change rapidly with the change in $\tau$, which is usually the case in real problems.
Assumption A2 requires that the true conditional mode does not lie in an extreme position.
Note that this assumption is valid even if the conditional density is highly skewed as long as its mode is not at the edge of its distribution.
Assumption A4 requires that the maximum of the conditional density is well separated, i.e., the maximum does not lie in a flat region.
Assumption A5 requires the conditional density to be upper-bounded by a polynomially decaying function.
With this assumption, we can ensure that the conditional density estimator $\hat{f}$ converges with a nearly uniform rate~\cite{blevins2015non}.

\subsection{Convergence of the Conditional Density Estimator $\hat{f}$}

First, we show that the conditional density estimator $\hat{f}$ converges to the true conditional density $f$.
The next theorem follows from Lemma~\ref{lem:quantile}.
\begin{theorem}[Uniform Convergence of $\hat{f}$]
	Under Assumptions A0 and A1, for a compact set $\mathcal{T} \subset (0,1)$,
	\begin{align}
	&\sup_{\tau \in \mathcal{T}}\left|\hat{f}(Q_{Y|X}(\tau|x)|x) - f(Q_{Y|X}(\tau|x)|x) \right| \nonumber \\
	&\hspace{6pt} = O_p\left(h^2  + \sqrt{\frac{1}{nh^2}}\right).
	\label{eq:convf}
	\end{align}
	\label{th:convf}
\end{theorem}
\paragraph{Proof Overview}
With some algebra, we obtain
\begin{align*}
&\left|\hat{f}(Q_{Y|X}(\tau|x)|x)) -f(Q_{Y|X}(\tau|x)|x)) \right|\\
& = \left( \hat{f}(Q_{Y|X}(\tau|x)|x)\cdot f(Q_{Y|X}(\tau|x)|x) \right)\\
& \hspace{12pt} \times\left| \hat{Q}_{Y|X}'(\tau|x)-Q_{Y|X}'(\tau|x) \right| ,
\end{align*}
where $\hat{Q}_{Y|X}'(\tau|x)=1/\hat{f}(Q_{Y|X}(\tau|x)|x)$ and  $Q_{Y|X}'(\tau|x)=1/f(Q_{Y|X}(\tau|x)|x)$.
By using the Taylor expansion theorem, we can upper-bound the term $\left| \hat{Q}_{Y|X}'(\tau|x)-Q_{Y|X}'(\tau|x) \right|$.
The uniform rate on $\mathcal{T}$ follows by applying Lemma~\ref{lem:quantile} to the upper-bound.


Note that, with additional smoothness assumptions on $Q_{Y|X}(\tau | x)$, the bias term in Eq.(\ref{eq:convf}) can be reduced from $h^2$ to $h^4$ leading to better convergence~\cite{belloni2016valid}. 

\subsection{Convergence of the Conditional Mode Estimator $\hat{m}(x)$}

Now, we turn to the convergence of the proposed mode estimator $\hat{m}(x)$.
Here, we show the consistencies of $\hat{\tau}_m(x)$ and $\hat{m}(x)$.
The next theorem is obtained via the asymptotic framework of an extremum estimation.
Applying  the uniform convergence rate of the conditional density estimator on $\mathcal{T}$ described in Theorem~\ref{th:convf}, we obtain the theorem below.
\begin{theorem}[Consistency of $\hat{\tau}_m(x)$] 
	Assume that Assumptions A0--A2 hold.
	In addition, assume that the bandwidth $h$ satisfies $h\to 0$, $nh^2\to \infty$ as $n\to \infty$.
	Then,
	\begin{align}
	\hat{\tau}_m(x) \cvp \tau_m(x).
	\end{align}
	\label{th:tau}
\end{theorem}
\paragraph{Proof Overview}
The theorem follows from the fact that $\hat{\tau}_m$ is defined as the extremum estimator maximizing $\hat{f}$.
We therefore prove the claim by using the consistency condition given by Newey and McFadden \cite{newey1994large}.
Specifically, we show that the consistency condition follows from the uniform convergence of the estimated conditional density $\hat{f}$ in Theorem~\ref{th:convf}.


The next theorem can also be derived by applying the uniform convergence rate of the conditional density estimator in Theorem~\ref{th:convf}.
\begin{theorem}[Consistency of $\hat{m}(x)$]
	Assume that Assumptions A0--A4 hold.
	In addition, assume that the bandwidth $h$ satisfies $h\to 0$, $nh^2\to \infty$ as $n\to \infty$.
	Then,
	\begin{align}
	\hat{m}(x) \cvp m(x).
	\end{align}
	\label{th:m}
\end{theorem}
\paragraph{Proof Overview}
We prove the theorem by evaluating the probability $P(|\hat{m}(x) - m(x)| \ge \varepsilon)$.
This probability can be bounded by the probability $P(|f(\hat{m}(x)|x) - f(m(x)|x)| \ge \delta)$ under Assumption A4.
We then bound $|f(\hat{m}(x)|x) - f(m(x)|x)|$ using the Lipschitz continuity of $f$, which shows that $P(|f(\hat{m}(x)|x) - f(m(x)|x)| \ge \delta)$ converges to zero.


Finally, we show the error rate of the proposed mode estimator $\hat{m}(x)$.
The next theorem can be obtained by applying Theorems~\ref{th:convf}, \ref{th:tau},  and \ref{th:m}.
\begin{theorem}[Error rate of $\hat{m}(x)$]
	Assume that Assumptions A0--A5 hold.
	In addition, assume that the bandwidth $h$ satisfies $h\to 0$ and $nh^{6-2k} \to \infty$ for sufficiently small $k>0$ as $n\to \infty$.
	Then,
	\begin{align}
	\hat{m}(x) - m(x) =  O_p\left( h^{(2-k)\gamma_2} + \frac{1}{\sqrt{n}} \right).
	\end{align}
	\label{th:error}
\end{theorem}
\paragraph{Proof Overview}
From the definitions of the  conditional mode and its estimator, we can represent $\hat{m}(x) - m(x)$ as $\{Q_{Y|X}(\hat{\tau}_m|x) -  Q_{Y|X}(\tau_m|x)  \} + \{ \hat{Q}_{Y|X}(\hat{\tau}_m|x)-Q_{Y|X}(\hat{\tau}_m|x) \}$.
The first term is evaluated by $\hat{\tau}_m - \tau_m$ using the Taylor expansion theorem, and  the convergence rate of $\hat{\tau}_m - \tau_m$ can be calculated under Assumption A5.
The convergence of the second term follows from Lemma~\ref{lem:quantile}.

\section{Comparisons of Error Rates to Existing Methods}
\label{sec:compare}

Theorem~\ref{th:error} shows that the bandwidth must satisfy $h\to 0$ and $nh^{6-2k} \to \infty$ to obtain an error rate $O_p\left( h^{(2-k)\gamma_2} + \frac{1}{\sqrt{n}} \right)$.
Here, we compare this result with existing modal regression methods.
In particular, we compare it with two representative studies of	 the linear modal regression~\cite{yao2014new,kemp2012quantile} and  nonparametric modal regression~\cite{chen2016nonparametric}.
\begin{result}[Linear Modal Regression~\cite{yao2014new,kemp2012quantile}]
	Under certain conditions, if $n\to \infty$, $h \to 0$, and $nh^5\to \infty$, then
	\begin{align}
	\hat{m}(x) - m(x) = O_p\left( h^2 + \frac{1}{\sqrt{nh^3}} \right).
	\end{align}
	\label{res:linear}
\end{result}
\begin{result}[Nonparametric Modal Regression~\cite{chen2016nonparametric}]
	Under certain conditions, if $n\to \infty$, $h \to 0$, and $nh^{p+5} / \log n \to \infty$, then
	\begin{align}
	\hat{m}(x) - m(x) = O_p\left( h^2 + \frac{1}{\sqrt{nh^{p+3}}} \right).
	\end{align}
	\label{res:nonparametric}
\end{result}

First, we observe that the proposed mode estimator (Theorem~\ref{th:error}) and the linear modal regression estimator (Result~\ref{res:linear}) have error rates that are not polynomial of the dimensionality $p$ of the input $x$.
Therefore, we expect these methods can perform well even in a high-dimensional setting by avoiding the curse of dimensionality.
Note that this desirable property relies heavily on the linearity assumptions of these models.
By contrast, the nonparametric estimator (Result~\ref{res:nonparametric}) does not require the linearity assumption and, therefore, it can be used even if the model is non-linear.
However, this method suffers from the curse of dimensionality because its error rate is dominated by the term $1 / \sqrt{nh^{p+3}}$, which converges very slowly when $p$ is large.

Second, we compare the proposed mode estimator and the linear modal regression estimator in detail, and show the advantage of the proposed estimator.
From Theorem~\ref{th:error}, we can see that the optimal bandwidth for the proposed mode estimator is $h=O_p(n^{-1/6})$.
Therefore, its optimal error rate is nearly $O_p(n^{-1/3})$ when $k$ is sufficiently small.
By contrast, Result~\ref{res:linear} implies that the optimal bandwidth for the kernel-based linear modal regression estimator is $h = O_p(n^{-1/7})$ and its optimal error rate is $O_p(n^{-2/7})$.
These results indicate that the proposed mode estimator is superior to the linear modal regression estimator because it has a faster convergence rate.
We note that the superiority of the proposed mode estimator comes from the fact that the proposed method requires stronger assumptions on the data.
That is, the proposed method requires the true $\tau$-th quantile function to be linear for all $\tau \in (0, 1)$ as in Assumption~A0, which is not required by the linear modal regression~\cite{yao2014new,kemp2012quantile}.

\section{Some Extensions of the Proposed Method}
\label{sec:extension}

Here, we describe two possible extensions of the proposed method.
The first extension is a scalable method for large datasets with an approximation.
The second extension is a sparse estimation method for high-dimensional data.

\subsection{Scalable Computation with an Approximation}

Algorithm~\ref{alg:proposed} can be computationally expensive if the candidate set $T$ is large because we need to solve the quantile regression, Eq.\ (\ref{eq:quantile}), $O(|T|)$ times.

Here, we propose a simple approximation algorithm based on the method of Volgushev et al.~\cite{volgushev2017distributed}.
In their method, we do not compute the coefficient $\hat{\beta}(\tau)$ for all $\tau \in \mathcal{T}$.
Instead, we compute $\hat{\beta}(\tau)$ only for a selected subset $\tau \in \mathcal{T}' \subsetneq \mathcal{T}$.
We then use the derived coefficients $\{\hat{\beta}(\tau')\}_{\tau' \in \mathcal{T}'}$ to approximate the remaining coefficients $\tilde{\beta}(\tau)$ for $\tau \in \mathcal{T} \setminus \mathcal{T}'$.
The approximate coefficient $\tilde{\beta}(\tau)$ can be derived by solving the following least squares regression:
\begin{align}
\tilde{\alpha} := \argmin_{\alpha \in \R^L} \sum_{\tau' \in T} \left\| \hat{\beta}(\tau') - \textstyle \sum_{\ell=1}^L \alpha_\ell \varphi_\ell(\tau') \right\|^2 ,
\label{eq:approx}
\end{align}
where $\{\varphi_\ell\}_{\ell=1}^L$ is a set of basis functions.
The approximate coefficient $\tilde{\beta}(\tau)$ is then given as $\tilde{\beta}(\tau) := \sum_{\ell=1}^L \tilde{\alpha}_\ell \varphi_\ell(\tau)$.
Because solving the least squares regression, Eq.\ (\ref{eq:approx}) is computationally far cheaper than solving the quantile regression several times, we can make Algorithm~\ref{alg:proposed} computationally more efficient using this approximation technique.

\subsection{Sparse and High-dimensional Modal Regression}

In the proposed method, we considered the ordinary quantile regression, Eq.\ (\ref{eq:quantile}).
Here, we consider the following $\ell_1$-regularized sparse quantile regression~\cite{belloni2011penalized}:
\begin{align}
\hat{\beta}(\tau) = \argmin_{b \in \R^p} & \sum_{i=1}^{n} \rho_\tau (y_i-x_i^\top b) \nonumber \\
& + \lambda \sqrt{\tau(1-\tau)} \sum_{j=1}^{p} \hat{\sigma}_j |b_j|,
\label{eq:sparse_quantile}
\end{align}
where $\lambda$ is a regularization parameter, and $\hat{\sigma}_j := \frac{1}{n} \sum_{i=1}^{n}x_{ij}^2$.
Note that Eq.\ (\ref{eq:sparse_quantile}) can be solved using linear programming also.

The advantage of the $\ell_1$-regularized formulation is that it can be applied even in the very high-dimensional setting where $p \gg n$.
We use this advantage of the $\ell_1$-regularized formulation, and extend Algorithm~\ref{alg:proposed} to a very high-dimensional setting.
In particular, we replace the quantile regression (\ref{eq:quantile}) in Algorithm~\ref{alg:proposed} with the $\ell_1$-regularized one, Eq.(\ref{eq:sparse_quantile}).

Our theoretical results can also be extended to the $\ell_1$-regularized setting.
The next theorem constitutes the basis of our theoretical results.

\begin{theorem}[Belloni and Chernozhukov~\cite{belloni2011penalized}]
	Under a suitable choice of $\lambda$, 
	\begin{align}
	\sup_{\tau \in \mathcal{T}} \left\| \hat{\beta}(\tau) - \beta(\tau) \right\| = O_p\left(\sqrt{\frac{s\log(n \vee p)}{n}}\right) 
	\end{align}
	holds with high probability, where $s$ is a parameter satisfying $\sup_{\tau \in \mathcal{T}} ||\beta(\tau)||_0 \leq s$.
\end{theorem}

Applying their result, we obtain the consistency of the conditional  mode estimator at a high-dimension.
\begin{theorem}[Consistency of $\hat{m}(x)$] 
	Assume that Assumptions A0--A4 hold. Further assume that the regularity condition in Belloni and Chernozhukov~\cite{belloni2011penalized} holds. 
	In addition, assume that the bandwidth $h$ satisfies $h\to 0$ and $nh^2/(s\log (p \vee n))\to \infty$ as $n\to \infty$.
	Then,
	\begin{align}
	\hat{m}(x) \cvp m(x).
	\end{align}
\end{theorem}
The theorem can be proved in the similar manner as Theorem~\ref{th:m} by using Theorem 5.

\section{Experiments}
\label{sec:experiment}

In this section, we show the efficacy of the proposed method via synthetic and real-world data experiments.

In the experiments, we compare the performances of the proposed method and the linear modal regression~\cite{yao2014new,kemp2012quantile}.
Note that the discussions in Section~\ref{sec:compare} indicated that the proposed method has a better convergence rate.
Therefore, we expect the proposed method to perform better than the linear modal regression.
We note that, we omitted the nonparametric modal regression~\cite{chen2016nonparametric} form the comparison except for the real-world data experiment.
This was because it performed significantly worse than the other two methods especially for $p \ge 3$ due to the curse of dimensionality.

We implemented the proposed method (Algorithm~\ref{alg:proposed}) using \texttt{R}.
To solve the quantile regression, we used the \texttt{quantreg} package, which was able to solve the problem in a reasonable time.
We implemented the linear modal regression using \texttt{Python}.

In the proposed method, 1,000 candidates of $\tau$ were selected uniformly from $[0.2, 0.8]$.
We set the bandwidth $h$ following the method of Koenker and Machado~\cite{koenker1999goodness}, as follows.
\begin{align}
h = n^{-1/6} z_\alpha^{2/3} \left\{1.5 \frac{\phi(\Phi^{-1}(\tau))}{2\Phi^{-1}(\tau)^2 +1} \right\}^{1/3}, 
\end{align}
where $\phi$ and $\Phi$ are the density function and the distribution function of the standard normal distribution, respectively.
$z_\alpha$ is a parameter that satisfies $\Phi(z_\alpha)= 1-\alpha/2$. 
We set $\alpha = 0.95$ in all the experiments.
Note that this bandwidth satisfies the conditions of Theorems \ref{th:tau}, \ref{th:m}, and \ref{th:error}. 
The bandwidth of the linear modal regression is chosen via cross validation.

\subsection{Synthetic Experiments}

We generated synthetic data as follows.
First, we generated the input data by $X_i \sim {\rm U}([0, 1]^p)$.
Then, we generated the output data by $Y_i = 1 + (1 + \nu \varepsilon) \sum_{j=1}^p X_{ij}$, where $\nu \sim {\rm Ga}(3, 1/2)$ and $\varepsilon \sim {\rm N}(1, 0.5^2)$.
Note that this function has a skewed noise distribution due to the gamma distribution.
Moreover, the noise distribution changes with the value of $X$.
Because the mode of the gamma distribution ${\rm Ga}(3, 1/2)$ is $1$, the modal function is ${\rm Mode}(Y|X)= 1+ 2\sum_{j=1}^p X_j$. 
Note that this model is a special case of \textit{the location-shift model}~\cite{koenker2005quantile} studied in the quantile regression literature.

We evaluated the proposed method in two settings.
The first setting was ``variable $p$ and fixed $n$'', and the second setting was ``fixed $p$ and variable $n$''.
In the first setting, we show that the proposed method scales well to high-dimensional settings. 
In the second setting, we confirm our theoretical result that the proposed method has a fast convergence rate.

The results for the ``variable $p$ and fixed $n$'' setting are shown in \figurename~1.
In the experiments, we varied the value of $p$ over $p = 1, 2, 3, 4,10, 20$, and $30$, while fixing $n$ to be $n = 500, 1,000$, and $2,000$.
To evaluate the performance of the modal regression methods, we randomly sampled 300 test points in the input space, and evaluated the mean square error ${\rm MSE} := \frac{1}{300} \sum_{x: \text{test points}} \left( \hat{m}(x) - m(x) \right)^2$.
The figures show the average MSE and its standard deviations over ten random data realizations.
We can observe the clear advantage of the proposed method over the linear modal regression.
That is, the proposed method attained smaller average MSE for all the cases.
We observed that the linear modal regression tended to stack in local optima while solving the non-convex optimization problem.
The proposed method does not have such local optimality issues, and hence its estimator was computationally very stable.

The results for the ``fixed $p$ and variable $n$'' setting are shown in \figurename~2.
The figure again shows the clear advantage of the proposed method in that it obtains smaller MSE.
This result also confirms our theoretical analysis that the proposed method has a fast convergence rate (see Section~\ref{sec:compare}).

\subsection{Synthetic Experiment in High-dimensional Setting}

We observed the performance of the proposed method in the high-dimensional setting where $p \gg n$ (see Section~\ref{sec:compare}).
As the baseline method, we adopted the linear modal regression~\cite{yao2014new,kemp2012quantile} with an additional $\ell_1$-regularization term that enforces the sparsity to its estimated regression coefficient.

In the experiment, we set $p=500$ and $n=100$.
We generated the synthetic data as follows.
We first generated the input data by $X_i \sim {\rm U}([0, 1]^{500})$.
We then generated the output data by $Y_i = 1 + (2 + \nu \varepsilon) \sum_{j=1}^5 X_{ij}$, where $\nu \sim {\rm Ga}(3, 1/2)$ and $\varepsilon \sim {\rm N}(1, 0.5^2)$.
This means that the first five elements of $X$ have effect on $Y$ and the others have no effects.
Because the mode of the Gamma distribution ${\rm Ga}(3, 1/2)$ is $1$, the modal function is ${\rm Mode}(Y|X)= 1+ 3\sum_{j=1}^5 X_j$.

\tablename~\ref{tab:highdim} shows the result.
In the experiment, we selected the regularization parameters of the proposed method and the linear modal regression using cross validation.
The table shows that the proposed method attained the smaller average MSE, which again confirms the effectiveness of the proposed method.

\subsection{Real-world Data Experiments}

We applied the proposed method to the speedflow diagrams used in Einbeck and Tutz~\cite{einbeck2006modelling}.
We obtained the data from the \texttt{hdrcde} package in \texttt{R}, where the task is to predict the speed from the flow.
In the experiment, we adopted the linear modal regression and the nonparametric modal regression as the baseline methods.
We note that this problem is low-dimensional, and thus the nonparametric modal regression could avoid the curse of dimensionality.

\figurename~3 shows the results on the proposed method, the linear modal regression, and the nonparametric modal regression.
We can observe that the estimated mode of the linear modal regression was biased toward downside when the flow is large, as shown in \figurename~3(b).
We conjecture that this was because the linear modal regression was affected by the fat tail of the conditional density.
By contrast, \figurename~3(a) shows that the proposed method well captured the mode of the conditional density.
This is because that the proposed method estimates the conditional density independently for each $\tau$, which is robust against the skewness of the distribution.
The result of the nonparametric modal regression in \figurename~3(c) shows that the estimated modes were biased toward downside.
We conjecture that this was because the nonparametric modal regression stack in local optima when searching for the maximum of the estimated conditional density.

\begin{table}[tb]
\centering
	\caption{Result for the $p \gg n$ experiment, where $p=500$ and $n=100$. Average MSE and its standard deviations over ten random data realizations are shown.}
	\label{tab:highdim}
	\begin{tabular}{cc}
		& Average MSE \\
		\hline
		Proposed & 0.015 $\pm$ 0.004 \\
		Linear Modal Regression & 0.018 $\pm$ 0.006
   \end{tabular}
\end{table}


\section{Conclusion}
\label{sec:conclusion}

In this paper, we proposed a new method for modal regression using multiple quantile regressions.
The proposed method has two advantages in that it is computationally stable because it has no initial parameter dependencies, and it is statistically efficient with a fast convergence rate.
Synthetic and real-world data experiments demonstrated the better performance of the proposed method compared to other existing methods.

\bibliographystyle{abbrv}
\bibliography{modal}

\newpage

\appendix
\section*{Appendix}

\section*{Proofs}\label{app:proofs}

\subsection*{Proof of Theorem 4.1}
Our proof is similar to that of Belloni, et al., \cite{belloni2016valid}.
\begin{proof}
	Let $\hat{Q}_{Y|X}(\cdot|x)= x'\hat{\beta}(\cdot)$. Using a Taylor expansion theorem for $Q_{Y|X}(\cdot|x)$ at $\tau + h, \tau-h$, we obtain
	\begin{align*}
	&\left|\hat{Q}_{Y|X}(\tau|x) - Q_{Y|X}(\tau|x) \right|\\
	& \leq \frac{|\hat{Q}_{Y|X}(\tau+h|x) - Q_{Y|X}(\tau+h|x) | }{h}\\
	& \hspace{12pt} +\frac{|\hat{Q}_{Y|X}(\tau-h|x) - Q_{Y|X}(\tau-h|x) | }{h}\\
	& \hspace{12pt} +\sup_{|\bar{\tau}-\tau| \leq h }|Q'''_{Y|X}(\bar{\tau}|x)|\cdot h^2\\
	&:= S_n(\tau)/h + Ch^2 ,
	\end{align*}
	where
	\begin{align*}
	S_n(\tau) =& \frac{|\hat{Q}_{Y|X}(\tau+h|x) - Q_{Y|X}(\tau+h|x) | }{h} \\
	& + \frac{|\hat{Q}_{Y|X}(\tau-h|x) - Q_{Y|X}(\tau-h|x) | }{h} , \\
	C =& \sup_{|\bar{\tau}-\tau| \leq h }|Q'''_{Y|X}(\bar{\tau}|x)| .
	\end{align*}
	Let $\hat{Q}_{Y|X}'(\tau|x)=1/\hat{f}(Q_{Y|X}(\tau|x)|x)$ and  $Q_{Y|X}'(\tau|x)=1/f(Q_{Y|X}(\tau|x)|x)$.
	Then, we have
	\begin{align*}
	&\left|\hat{f}(Q_{Y|X}(\tau|x)|x)) -f(Q_{Y|X}(\tau|x)|x)) \right|\\	& =\frac{\left| \hat{Q}_{Y|X}'(\tau|x)-Q_{Y|X}'(\tau|x)\right|}{\left( \hat{Q}_{Y|X}'(\tau|x)\cdot Q_{Y|X}(\tau|x)\right)}\\
	& = \left( \hat{f}(Q_{Y|X}(\tau|x)|x)\cdot f(Q_{Y|X}(\tau|x)|x) \right)\\
	& \hspace{12pt} \times\left| \hat{Q}_{Y|X}'(\tau|x)-Q_{Y|X}'(\tau|x) \right|\\
	&\leq \left( \hat{f}(Q_{Y|X}(\tau|x)|x)\cdot f(Q_{Y|X}(\tau|x)|x) \right) \\
	& \hspace{12pt} \times (S_n(\tau)/h + Ch^2).
	\end{align*}
	This implies $\hat{f}(1- S_n(\tau)/h - Ch^2) \leq  f$.
	Recall that $f$ is bounded and $S_n(\tau)/h$ and $Ch^2$ go to 0 according to Lemma~\ref{lem:quantile}.
	This implies that $\hat{f}$ is bounded, and hence $|\hat{f}-f|=O_p(S_n(\tau)/h + h^2)$.
	The uniform rate on $\mathcal{T}$ immediately follows from this result.
\end{proof}

\subsection*{Proof of Theorem 4.2}

\begin{proof}
	$\hat{\tau}_m(x)$ is defined as an extremum estimator, i.e., as the maximizer of the objective function $\hat{f}(Q_{Y|X}(\tau|x)|x))$.
	To prove the consistency of extremum estimators, we invoke the lemma as below.
	\begin{lemma}[Sufficient conditions for the consistency of extremum estimators, Newey and McFadden (1994)\cite{newey1994large}]
		
		Let $\hat{\theta}_n$ be a maximizer of $M_n(\theta)$, where $M_n(\theta)$ is a random function.
		If there is a  function $M(\theta)$ such that
		\medskip
		
		(i) $M(\theta)$ is uniquely maximized at $\theta_0$,
		
		(ii) The parameter space $\Theta$ is compact,
		
		(iii) $M(\theta)$ is continuous, and
		
		(iv) $M_n(\theta)$ converges uniformly in probability to $M(\theta)$,
		\medskip
		
		then, $\hat{\theta}_n \cvp \theta_0.$
	\end{lemma}
	
	$M$, $M_n$, $\theta$, and $\theta_n$ correspond to $f$, $\hat{f}$, $\tau_m$ and $\hat{\tau}_m$, respectively in our case.
	Therefore, we will check the conditions (i)--(iv).
	The conditions (i), (ii), and (iii) are satisfied under Assumptions A2 and A3.
	Finally, we verify the condition (iv).
	From Theorem 1, if $nh^2 \to \infty$ and $h\to 0$ as $n \to \infty$, $\left|\hat{f}(Q_{Y|X}(\tau|x)|x) -f(Q_{Y|X}(\tau|x)|x)\right|$ converges to 0 in probability uniformly on $\mathcal{T}$.
	Therefore the extremum estimator $\hat{\tau}_m(x)$ converges in probability to $\tau_m(x)$.
\end{proof}

\subsection*{Proof of Theorem 4.3}

\begin{proof}
	From Assumption A4,
	\begin{align*}
	&P\left( |\hat{m}(x) - m(x)| \geq \varepsilon  \right) \\&\leq P\left( |f(\hat{m}(x)|x) - f(m(x)|x) |\geq \delta  \right),
	\end{align*}
	holds.
	Here, we have
	\begin{align*}
	&\left|f(m(x)|x) - f(\hat{m}(x)|x)\right|\\
	& = \left|f(Q_{Y|X}(\tau_m|x)|x) - f(\hat{Q}_{Y|X}(\hat{\tau}_m|x)|x)\right| \\
	& \leq   4 \sup_{\tau \in \mathcal{T}}\left|f(Q_{Y|X}(\tau|x)|x) - \hat{f}(Q_{Y|X}(\tau|x)|x)\right|\\
	&\hspace{12pt} + L_x  \sup_{\tau \in \mathcal{T}} ||\hat{\beta}(\tau)-\beta(\tau)||\\
	&= O_p\left(h^2  + \sqrt{\frac{1}{nh^2}}\right) + O_p\left(\sqrt{\frac{1}{n}}\right).
	\end{align*}
	Therefore, if $nh^2 \to \infty$ and $h\to 0$ as $n \to \infty$, $\hat{m}(x)$ converges in probability to $m(x)$.
\end{proof}

\subsection*{Proof of Theorem 4.4}

\begin{proof}
	From the definition of the conditional mode, we have
	\begin{align*}
	&\hat{m}(x) - m(x) \\
	& = \hat{Q}_{Y|X}(\hat{\tau}_m|x)-Q_{Y|X}(\tau_m|x)\\
	& =  \left\{Q_{Y|X}(\hat{\tau}_m|x) -  Q_{Y|X}(\tau_m|x)  \right\} \\
	&\hspace{12pt} + \left\{ \hat{Q}_{Y|X}(\hat{\tau}_m|x)-Q_{Y|X}(\hat{\tau}_m|x) \right\}\\
	&=\frac{\hat{\tau}_m - \tau_m}{f(Q_{Y|X}(\bar{\tau}_m|x))} + O_p\left(\sqrt{\frac{1}{n}}\right).
	\end{align*}
	Because $f$ and $Q_{Y|X}$ are continuous, $f(Q_{Y|X}(\bar{\tau}_m|x)) \cvp f(Q_{Y|X}(\tau_m|x))$ holds from the continuous mapping theorem.
	
	Next we evaluate $|\hat{\tau}_m - \tau_m|$ under Assumption A5. 
	Let $\varepsilon' \in (0,1)$ be a given parameter and let $\delta', K, \gamma_1, \gamma_2, K_{\varepsilon'}$, and $n_{\varepsilon'}$ satisfy Assumption A5.
	Define $a_n := (h^2  + 1/\sqrt{nh^2})^{-1}$ and let $b_n$ be a non-negative sequence that satisfies $a_n b_n \to \infty$. 
	We then define
	\begin{align*}
	d_n := \left( \frac{K_1K_{\varepsilon'} \vee  a_n b_n}{a_n K_1}  \right)^{1/\gamma_1}. 
	\end{align*}
	
	We now show, that with a probability of at least $1-\varepsilon'$, there exists an $N \geq n_{\varepsilon'}$ such that for all $n \geq N$, the following (a), (b), and (c), are true:
	\begin{enumerate}
		\item[(a)] $d_n\geq (K_{\varepsilon'}/a_n)^{\gamma_2}$,
		\item[(b)] $d_n \leq \delta'$,
		\item[(c)] $\sup_{\tau \in \mathcal{T}}| \hat{f}(Q_{Y|X}(\tau|x)|x)) -f(Q_{Y|X}(\tau|x)|x)) | \leq b_n$.
	\end{enumerate}
	\paragraph{(a) is true:}
	For sufficiently large $n$, 
	\[ d_n^{1/\gamma_2} \geq \left( \frac{K_1K_{\varepsilon'} \vee  a_n b_n}{a_n K_1}  \right)^{1/(\gamma_1\gamma_2)} \geq \left( \frac{K_{\varepsilon'}}{a_n} \right) \] holds with with a probability of at least $1-\varepsilon'$ because $\gamma_1 \gamma_2 \geq 1$ and $a_n b_n \to \infty$. 
	\paragraph{(b) is true:}
	It follows from the fact that $d_n = o_p(1)$ and $\delta'>0$.
	\paragraph{(c) is true:}
	From Theorem 4.1,
	
	\begin{align*}
	\sup_{\tau \in \mathcal{T}}| \hat{f}(Q_{Y|X}(\tau|x)|x)) -f(Q_{Y|X}(\tau|x)|x) |= O_p(1/a_n).
	\end{align*}
	Using the condition $a_n b_n \to \infty$, we can conclude that  $\sup_{\tau \in \mathcal{T}}| \hat{f}(Q_{Y|X}(\tau|x)|x) -f(Q_{Y|X}(\tau|x)|x) | \leq  b_n$. 
	
	We now prove the theorem by using (a), (b), and (c).
	Let 
	\[\mathcal{T}_{d_n} := \{ \tau \in \mathcal{T}: |\tau - \tau_m| \leq d_n\}. \]
	Then, the following inequality is true for all $n \geq n_{\varepsilon'}$ with a probability of  at least $1-\varepsilon'$.
	\begin{align*}
	&\sup_{\tau \in \mathcal{T}\setminus \mathcal{T}_{d_n}} \hat{f}(Q_{Y|X}(\tau|x)|x) \\& <\sup_{\tau \in \mathcal{T}}f(Q_{Y|X}(\tau|x)|x)- K_1(d_n \wedge \delta')^{\gamma_1}\\
	&\leq \sup_{\tau \in \mathcal{T}}f(Q_{Y|X}(\tau|x)|x)-K_1d_n^{\gamma_1}\\&\leq \sup_{\tau \in \mathcal{T}}f(Q_{Y|X}(\tau|x)|x)-b_n\\
	& \leq \sup_{\tau \in \mathcal{T}}\hat{f}(Q_{Y|X}(\tau|x)|x)\\
	&=\hat{f}(Q_{Y|X}(\hat{\tau}_m|x)|x),
	\end{align*}
	where we used (a), (b), and (c) and the definition of the conditional mode estimator.
	This relation implies that $\hat{\tau}_m \subset \mathcal{T}_{d_n}$ holds with probability at least $1-\varepsilon'$ for $n \geq n_{\varepsilon'}$.
	
	The first inequality holds by Assumption A4 and A5, and condition (a) under which $|\tau - \tau_m| > d_n \geq (K_{\varepsilon'}/a_n)^{\gamma_2}$. The second inequality is a direct result of (b). The third inequality holds from the definition of $d_n$. The fourth inequality follows from (c). Finally, the fifth equality follows by the definition of $\hat{\tau}_m$. 
	
	Therefore, for all $n \geq n_{\varepsilon'}$, with probability at least $1-\varepsilon'$, $|\hat{\tau}_m -\tau_m| \leq d_n$ holds, and subsequently 
	$|\hat{\tau}_m -\tau_m| = O_p(b_n^{\gamma_2})$ follows.
	
	By setting $b_n=h^{(2-k)}$, we obtain $a_nb_n \to \infty$ by recalling that $nh^{6-2k} \to \infty$.
	A fast rate can be achieved when $k>0$ is sufficiently small.
\end{proof}

\subsection*{Proof of Theorem 4.6}
\begin{proof}
	
	First, we have the convergence rate of the conditional density estimator as follows.
	Applying Theorem 4.5 to the evaluation of $S_n(\tau)$ in the proof of Theorem 4.1, we obtain
	\begin{align}
	&\sup_{\tau \in \mathcal{T}}\left|\hat{f}(Q_{Y|X}(\tau|x)|x) - f(Q_{Y|X}(\tau|x)|x) \right| \nonumber \\
	&\hspace{6pt} = O_p\left(h^2  + \sqrt{\frac{s\log (n \vee p)}{nh^2}}\right).
	\end{align}
	Next, using the uniform convergence result, we derive $\hat{\tau}_m \cvp \tau_m$ and $\hat{m}(x) \cvp m(x)$ under the bandwidth assumption: $h$ satisfies $h\to 0$ and $nh^2/(s\log (p \vee n))\to \infty$ as $n\to \infty$.
	
\end{proof}

\section*{Figures}
\label{sec:fig}

\begin{figure}[h]
\begin{tikzpicture}
\begin{semilogxaxis}
[
	scale=1.0,
	xlabel={Dimensionality $p$},
	ylabel={Average MSE},
	ymin=0,
	ymax=0.04,
	scaled ticks=false,
	yticklabel style={
		/pgf/number format/fixed,
		/pgf/number format/precision=3,
		/pgf/number format/fixed zerofill
	},
	width=0.99\textwidth,
	height=5.5cm,
	legend style={font=\normalsize},
	legend entries={Proposed, Linear}
]
\addplot [mark=*, mark size=3, color=blue, error bars/.cd, y dir=both, y explicit] table [x=dim, y=prop, y error=propsd, col sep=comma] {n_500.csv};
\addplot [mark=square*, mark size=3, color=red, error bars/.cd, y dir=both, y explicit] table [x=dim, y=yaoli, y error=yaolisd, col sep=comma] {n_500.csv};
\end{semilogxaxis}
\end{tikzpicture}
\subcaption{$n=500$}
\label{fig:n500}

\hfill

\begin{tikzpicture}
\begin{semilogxaxis}
[
	scale=1.0,
	xlabel={Dimensionality $p$},
	ylabel={Average MSE},
	ymin=0,
	ymax=0.020,
	scaled ticks=false,
	yticklabel style={
		/pgf/number format/fixed,
		/pgf/number format/precision=3,
		/pgf/number format/fixed zerofill
	},
	width=0.99\textwidth,
	height=5.5cm,
]
\addplot [mark=*, mark size=3, color=blue, error bars/.cd, y dir=both, y explicit] table [x=dim, y=prop, y error=propsd, col sep=comma] {n_1000.csv};
\addplot [mark=square*, mark size=3, color=red, error bars/.cd, y dir=both, y explicit] table [x=dim, y=yaoli, y error=yaolisd, col sep=comma] {n_1000.csv};
\end{semilogxaxis}
\end{tikzpicture}
\subcaption{$n=1,000$}
\label{fig:n1000}

\hfill

\begin{tikzpicture}
\begin{semilogxaxis}
[
	scale=1.0,
	xlabel={Dimensionality $p$},
	ylabel={Average MSE},
	ymin=0,
	ymax=0.020,
	scaled ticks=false,
	yticklabel style={
		/pgf/number format/fixed,
		/pgf/number format/precision=3,
		/pgf/number format/fixed zerofill
	},
	width=0.99\textwidth,
	height=5.5cm,
]
\addplot [mark=*, mark size=3, color=blue, error bars/.cd, y dir=both, y explicit] table [x=dim, y=prop, y error=propsd, col sep=comma] {n_2000.csv};
\addplot [mark=square*, mark size=3, color=red, error bars/.cd, y dir=both, y explicit] table [x=dim, y=yaoli, y error=yaolisd, col sep=comma] {n_2000.csv};
\end{semilogxaxis}
\end{tikzpicture}
\subcaption{$n=2,000$}
\label{fig:n2000}
\caption*{Figure 1: Results for the ``variable $p$ and fixed $n$'' experiment. The proposed method and the linear modal regression are compared for $p=1,2,3,4,10,20$, and $30$ and $n=500, 1000$, and $2000$ over ten random data realizations. The average MSE and its standard deviations are plotted.}
\label{fig:variablep}
\end{figure}
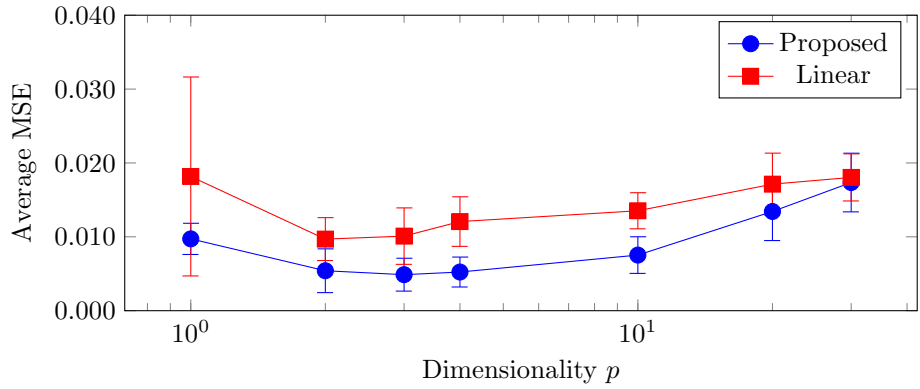
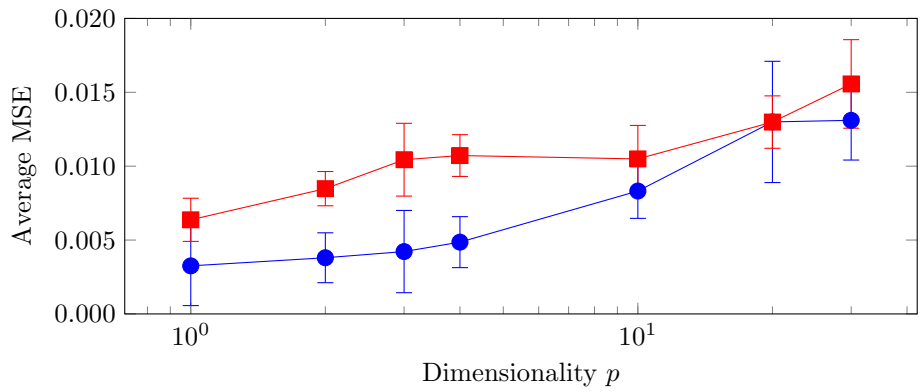
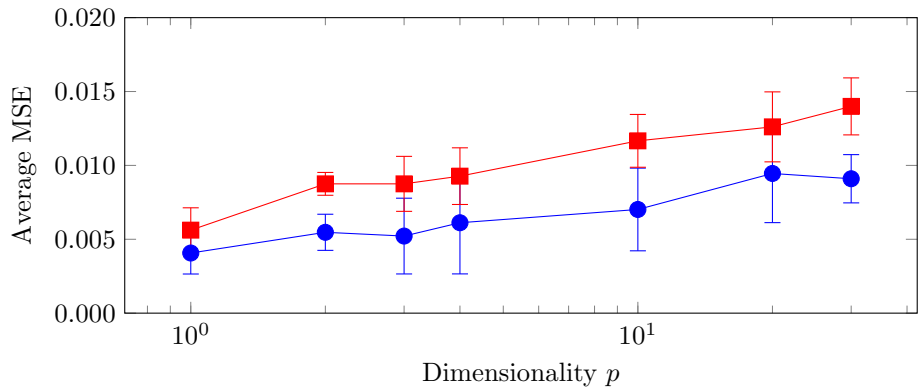

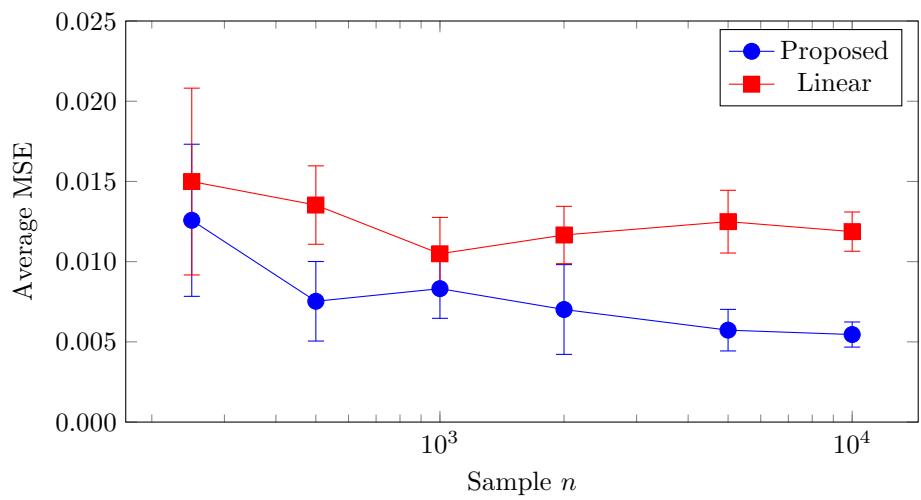
\begin{figure}[t]
\begin{tikzpicture}
\begin{semilogxaxis}
[
	scale=1.0,
	xlabel={Sample $n$},
	ylabel={Average MSE},
	ymin=0,
	ymax=0.025,
	scaled ticks=false,
	yticklabel style={
		/pgf/number format/fixed,
		/pgf/number format/precision=3,
		/pgf/number format/fixed zerofill
	},
	width=0.99\textwidth,
	height=6.9cm,
	legend style={font=\normalsize},
	legend entries={Proposed, Linear}
]
\addplot [mark=*, mark size=3, color=blue, error bars/.cd, y dir=both, y explicit] table [x=sample, y=prop, y error=propsd, col sep=comma] {p_10.csv};
\addplot [mark=square*, mark size=3, color=red, error bars/.cd, y dir=both, y explicit] table [x=sample, y=yaoli, y error=yaolisd, col sep=comma] {p_10.csv};
\end{semilogxaxis}
\end{tikzpicture}
	\caption{Result for the ``fixed $p$ and variable $n$'' experiment with $p=10$ over ten random data realizations. The average MSE and its standard deviations are plotted.}
	\label{fig:variablen}
\end{figure}
\thispagestyle{empty}

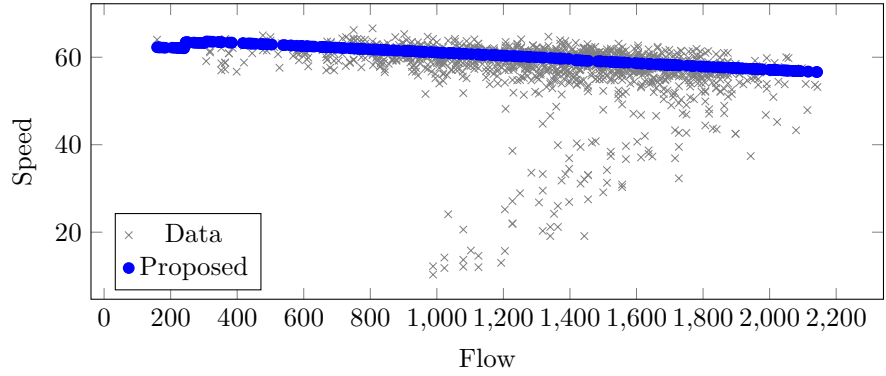
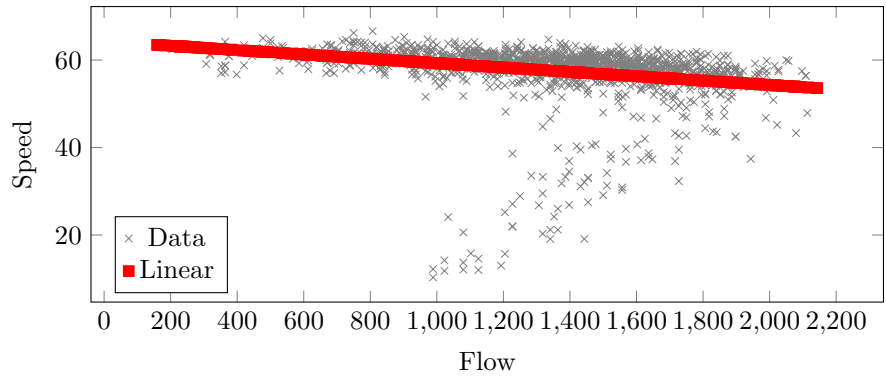
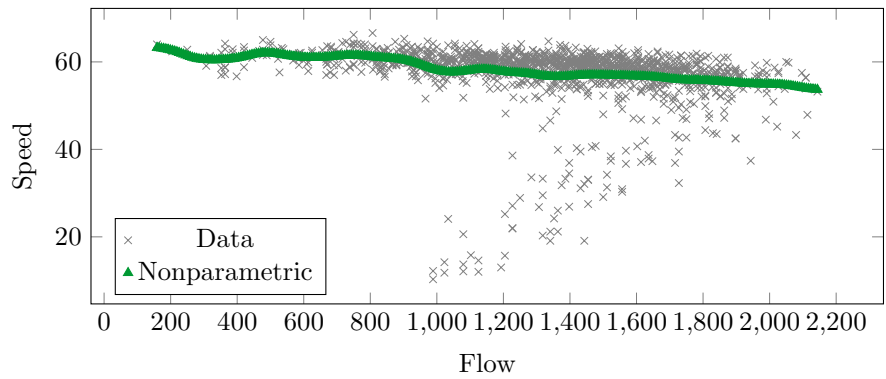
\begin{figure}[t]

\begin{tikzpicture}
\begin{axis}
[
	scale=1.0,
	xlabel={Flow},
	ylabel={Speed},
	scaled ticks=false,
	tick label style={/pgf/number format/fixed},
	width=0.99\textwidth,
	height=5.5cm,
	legend style={font=\normalsize},
	legend entries={Data,Proposed},
	legend pos=south west
]
\addplot [only marks, mark=x, mark size=2, ultra thin, color={rgb:black,2;white,2}] table [x=flow, y=speed, col sep=comma] {lane2.csv};
\addplot [only marks, mark=*, mark size=2, color=blue] table [x=x_pred, y=y_prop, col sep=comma] {lane2.csv};
\end{axis}
\end{tikzpicture}
\subcaption{Proposed}
\label{fig:lane2proposed}

\hfill

\begin{tikzpicture}
\begin{axis}
[
	scale=1.0,
	xlabel={Flow},
	ylabel={Speed},
	scaled ticks=false,
	tick label style={/pgf/number format/fixed},
	width=0.99\textwidth,
	height=5.5cm,
	legend style={font=\normalsize},
	legend entries={Data, Linear},
	legend pos=south west
]
\addplot [only marks, mark=x, mark size=2, ultra thin, color={rgb:black,2;white,2}] table [x=flow, y=speed, col sep=comma] {lane2.csv};
\addplot [only marks, mark=square*, mark size=2, color=red] table [x=x_pred, y=y_yaoli, col sep=comma] {lane2.csv};
\end{axis}
\end{tikzpicture}
\subcaption{Linear Modal Regression}
\label{fig:lane2linear}

\hfill

\begin{tikzpicture}
\begin{axis}
[
	scale=1.0,
	xlabel={Flow},
	ylabel={Speed},
	scaled ticks=false,
	tick label style={/pgf/number format/fixed},
	width=0.99\textwidth,
	height=5.5cm,
	legend style={font=\normalsize},
	legend entries={Data, Nonparametric},
	legend pos=south west
]
\addplot [only marks, mark=x, mark size=2, ultra thin, color={rgb:black,2;white,2}] table [x=flow, y=speed, col sep=comma] {lane2.csv};
\addplot [only marks, mark=triangle*, mark size=2, color={rgb:green,3;blue,1;black,1}] table [x=xchen, y=ychen, col sep=comma] {chen.csv};
\end{axis}
\end{tikzpicture}
\subcaption{Nonparametric Modal Regression}
\label{fig:lane2nonp}

\caption*{Figure 3: Results for the speedflow data experiment: The estimated modes of the proposed method, the linear modal regression, and the nonparametric modal regression are plotted.}
\label{fig:speedflow}
\end{figure}

\end{document}